\documentclass[11pt,a4paper]{article}
\usepackage[hyperref]{naaclhlt2019}
\usepackage{times}
\usepackage{latexsym}

\usepackage{url}
\usepackage{amsmath}
\usepackage{amssymb}
\usepackage{graphicx}
\usepackage{subcaption}
\usepackage{multirow}
\usepackage{xspace}

\DeclareMathOperator*{\softmax}{softmax}

\DeclareMathOperator*{\argmin}{argmin}

\DeclareMathOperator*{\ce}{CE}
\newcommand{\yspace}{\mathcal{Y}}
\newcommand{\w}{\mathbf{W}}
\newcommand{\info}{\mathit{inf}}
\newcommand{\vito}{\mathit{vit}}

\newcommand{\window}{n}

\newcommand{\blstmcrf}{BLSTM-CRF\xspace}
\newcommand{\blstmcnncrf}{BLSTM-CRF+\xspace}
\newcommand{\gd}{GD\xspace}

\aclfinalcopy 

\newcommand{\x}{\boldsymbol{x}}
\newcommand{\y}{\boldsymbol{y}}
\newcommand{\yv}{\mathbf{y}}
\newcommand{\relyspace}{\mathcal{Y}_{R}}
\newcommand{\relyspacer}{\mathcal{Y}_{R'}}
\newcommand{\ltok}{\ell_{\mathrm{token}}}

\newcommand{\infnet}{\mathbf{A}_{\Psi}}
\newcommand{\infnetnoparams}{\mathbf{A}}

\newcommand{\infnetBase}{\textrm{infnet}\xspace}
\newcommand{\infnetLarge}{\textrm{infnet+}\xspace}

\title{An Exploration of Inference Network Architectures\\for Sequence Labeling}
\title{Benchmarking Approximate Inference 
in Neural Structured Prediction}
\title{Benchmarking Approximate Inference Methods \\ for Neural Structured Prediction}

\author{Lifu Tu \ \ \ \ \ \ Kevin Gimpel \\
  Toyota Technological Institute at Chicago, Chicago, IL, 60637, USA \\
  {\tt \{lifu,kgimpel\}@ttic.edu}
 \\}

\date{}

\begin{document}
\maketitle
\begin{abstract}

Exact structured inference with neural network scoring functions is computationally challenging but several methods have been proposed for approximating inference. One approach is to perform gradient descent with respect to the output structure directly~\citep{belanger2016structured}. Another approach, proposed recently, is to train a neural network (an ``inference network'') to perform inference~\citep{tu-18}.  
In this paper, we compare these two families of inference methods on three sequence labeling datasets. We choose sequence labeling because it permits us to use exact inference as a benchmark in terms of speed, accuracy, and search error. Across datasets, we demonstrate that inference networks achieve a better speed/accuracy/search error trade-off than gradient descent, while also being faster than exact inference at similar accuracy levels. We find further benefit by combining inference networks and gradient descent, using the former to provide a warm start for the latter.\footnote{Code is available at 
\texttt{github.com/lifu-tu/} \texttt{Benchmarking}\texttt{ApproximateInference}}
\end{abstract}

\section{Introduction}

Structured prediction models commonly involve complex inference problems for which finding exact solutions is intractable~\citep{Cooper:1990:CCP:77754.77762}. There are generally two ways to address this difficulty. One is to restrict the model family to those for which inference is feasible. For example, state-of-the-art methods for sequence labeling use structured energies that decompose into label-pair potentials and then use rich neural network architectures to define the potentials~\citep[\emph{inter alia}]{2011:NLP,LampleBSKD16,ma-hovy:2016:P16-1}.
Exact dynamic programming algorithms like the Viterbi algorithm can be used for inference. 
\hyphenation{com-pu-ta-tionally-in-tractable}
The second approach is to retain computationally-intractable scoring functions but then use approximate methods for inference. 
For example, some researchers relax the structured output space from a discrete space to a continuous one 
and then use gradient descent to maximize the score function with respect to the output \citep{belanger2016structured}. 
Another approach is to train a neural network (an ``inference network'') to output a structure in the relaxed space that has high score under the structured scoring function~\citep{tu-18}. 
This idea was proposed as an alternative to gradient descent in the context of structured prediction energy networks~\citep{belanger2016structured}.

In this paper, we empirically compare exact inference, gradient descent, and inference networks for three sequence labeling tasks. 
We train conditional random fields (CRFs) for sequence labeling with neural networks used to define the potentials. We choose a scoring function that permits exact inference via Viterbi so that we can benchmark the approximate methods in terms of search error in addition to speed and accuracy. 
We consider three families of neural network architectures to serve 
as inference networks: convolutional neural networks (CNNs), recurrent neural networks (RNNs), and sequence-to-sequence models with attention (seq2seq; \citealp{sutskever2014sequence,BahdanauCB14}). We also use multi-task learning while training inference networks, combining the structured scoring function with a local cross entropy loss. 

Our empirical findings can be summarized as follows. 
Gradient descent works reasonably well for tasks with small label sets and primarily local structure, like part-of-speech tagging. However, gradient descent struggles on tasks with long-distance dependencies, 
even with small label set sizes. 
For tasks with large label set sizes, inference networks and Viterbi perform comparably, with Viterbi taking much longer. 
In this regime, it is difficult for gradient descent to find a good solution, even with many  iterations. 

In comparing inference network architectures, (1) CNNs are the best choice for tasks with primarily local structure, like part-of-speech tagging; (2) RNNs can handle  longer-distance dependencies while still offering high decoding speeds; and (3) seq2seq networks consistently work better than RNNs, but are also the most computationally expensive.

We also compare search error between gradient descent and inference networks and measure correlations with input likelihood. We find that inference networks achieve lower search error on instances with higher likelihood (under a pretrained language model), while for gradient descent the correlation between search error and likelihood is closer to zero. This shows the impact of the use of dataset-based learning of inference networks, i.e., they are more effective at amortizing inference for more common inputs. 

Finally, we experiment with two refinements of inference networks. 
The first fine-tunes the inference network parameters for a single test example to minimize the energy of its output. 
The second uses an inference network to provide a warm start for gradient descent. 
Both lead to reductions in search error and higher accuracies for certain tasks, with the warm start method leading to a better speed/accuracy trade-off.

\section{Sequence Models}

For sequence labeling tasks, given an input sequence $\x = \langle x_1, x_2,..., x_{|\x|}\rangle$, we wish to output a sequence $\y = \langle \yv_1, \yv_2,..., \yv_{|\x|}\rangle \in \yspace(\x)$. Here $\yspace(\x)$ is the structured output space for $\x$. Each label $\yv_t$ is represented as an $L$-dimensional one-hot vector 
where $L$ is the number of labels.

Conditional random fields (CRFs; \citealp{Lafferty:2001:CRF:645530.655813}) form one popular class of methods for structured prediction, especially for sequence labeling. 
We define our structured energy function to be similar to those often used in CRFs for sequence labeling:
\begin{align}
\smallskip
&E_{\Theta}(\x, \y)  = \nonumber 
\\ & -\left(\sum_{t} \sum_{i=1}^L y_{t,i} \left(\mathbf{u}_i^\top f(\x,t)\right) + \sum_{t} \yv_{t-1}^\top \mathbf{W} \yv_{t}\right) \nonumber
\end{align}
where $y_{t,i}$ is the $i$th entry of the vector $\yv_t$. In the standard discrete-label setting, each $\yv_t$ is a one-hot vector, but this energy is generalized to be able to use both discrete labels and continuous relaxations of the label space, which we will introduce below. 
Also, we use $f(\x,t)\in \mathbb{R}^d$ to denote the ``input feature vector'' for position $t$, $\mathbf{u}_i\in\mathbb{R}^d$ is a label-specific parameter vector used for modeling the local scoring function, and $\mathbf{W}\in\mathbb{R}^{L \times L}$ is a parameter matrix learned to model label transitions. For the feature vectors we use a bidirectional long short-term memory (BLSTM; \citealp{Hochreiter:1997:LSM:1246443.1246450}), so this forms a \blstmcrf~\citep{LampleBSKD16, ma-hovy:2016:P16-1}. 

For training, we use the standard conditional log-likelihood objective for CRFs, using the forward and backward dynamic programming algorithms to compute gradients. 
For a given input $\x$ at test time, prediction is done by choosing the output with the lowest energy:
\begin{align}
\argmin_{\y\in\yspace(\x)}E_{\Theta}(\x, \y) \nonumber
\end{align}
The Viterbi algorithm can be used to solve this problem exactly for the energy defined above. 

\subsection{Modeling Improvements: \blstmcnncrf}
\label{sec:strongerBaselines}

For our experimental comparison, we consider two CRF variants. The first is the basic model described above, which we refer to as \blstmcrf. Below we describe three additional techniques that we add to the basic model. 

We will refer to the CRF with these three techniques as \blstmcnncrf. 
Using these two models permits us to assess the impact of model complexity and performance level on the inference method comparison.

\paragraph{Word Embedding Fine-Tuning.} 
We used pretrained, fixed word embeddings when using the \blstmcrf model, but for the more complex \blstmcnncrf model, we fine-tune the pretrained word embeddings during training. 

\paragraph{Character-Based Embeddings.}
Character-based word embeddings provide consistent improvements in sequence labeling  \citep{LampleBSKD16,ma-hovy:2016:P16-1}. In addition to pretrained word embeddings, we produce a character-based embedding for each word using a character convolutional network like that of \citet{ma-hovy:2016:P16-1}. The filter size is 3 characters and the character embedding dimensionality is 30. We use max pooling over the character sequence in the word and the resulting embedding is concatenated with the word embedding before being passed to the BLSTM. 

\paragraph{Dropout.}
We also add dropout during training~\citep{dropout}. 
Dropout is applied before the character embeddings are fed into the CNNs, at the final word embedding layer before the input to the BLSTM, and after the BLSTM. The dropout rate is 0.5 for all experiments.

\section{Gradient Descent for Inference}
\label{sec:gd}
To use gradient descent (GD) for structured inference, researchers typically relax the output space from a discrete, combinatorial space to a continuous one and then use gradient descent to solve the following optimization problem: 
\begin{align}
\argmin_{\y\in\relyspace(\x)}E_\Theta(\x, \y) \nonumber  
\end{align}
where $\relyspace$ is the relaxed continuous output space. For sequence labeling, $\relyspace(\x)$ consists of length-$|\x|$ sequences of probability distributions over output labels. 
To obtain a discrete labeling for evaluation, the most probable label at each position is returned. 

There are multiple settings in which gradient descent has been used for structured inference, e.g., image generation \citep{Johnson2016Perceptual}, structured prediction energy networks \citep{belanger2016structured}, and machine translation \citep{DBLP:conf/emnlp/HoangHC17}. 
Gradient descent has the advantage of simplicity. Standard autodifferentiation toolkits can be used to compute gradients of the energy with respect to the output once the output space has been relaxed. 
However, one challenge is maintaining constraints on the variables being optimized. 

Therefore, we actually perform gradient descent in an even more relaxed output space $\relyspacer(\x)$ which consists of length-$|\x|$ sequences of 
vectors, where each vector $\yv_t\in\mathbb{R}^L$. 
When computing the energy, we use a softmax transformation on each $\yv_t$, solving the following optimization problem with gradient descent:
\begin{align}
\argmin_{{\y}\in\relyspacer(\x)}E_\Theta(\x, \softmax({\y})) 
\label{eq:gdinf}
\end{align}
where the softmax operation above is applied independently to each vector $\yv_t$ in the output structure $\y$.

\section{Inference Networks}
\label{sec:infnet}

\citet{tu-18} define an \textbf{inference network} (``infnet'') $\infnet : \mathcal{X}\rightarrow \relyspace$ 
and train it 
with the goal that
\begin{align}
\infnet(\x) \approx \argmin_{\y\in\relyspace(\x)}E_\Theta(\x, \y) \nonumber  
\end{align}
where $\relyspace$ is the relaxed continuous output space as defined in Section~\ref{sec:gd}. 

For sequence labeling, for example, an inference network $\infnet$ takes a sequence $\x$ as input and outputs a distribution over labels for each position in $\x$. Below we will consider three families of neural network architectures for $\infnet$.

For training the inference network parameters $\Psi$, \citet{tu-18} explored stabilization and regularization terms and found that a local cross entropy loss consistently worked well for sequence labeling. We use this local cross entropy loss in this paper, so we perform learning by solving the following:

\begin{align}
\argmin_{\Psi}\!\!\sum_{\langle\x,\y\rangle} \!\! E_\Theta(\x, \infnet(\x))\! +\! \lambda\ltok(\y, \infnet(\x))\nonumber
\end{align}

\noindent where 
the sum is over $\langle \x, \y\rangle$ pairs in the training set. 
%
The token-level loss is defined:  
\begin{align}
\ltok(\y, \infnetnoparams(\x)) = \sum_{t=1}^{|\y|} \ce (\yv_t, \infnetnoparams(\x)_t)  
\label{eq:token_loss}
\end{align}
where $\yv_t$ is the $L$-dimensional one-hot label vector at position $t$ in $\y$, $\infnetnoparams(\x)_t$ is the inference network's output distribution 
at position $t$, 
and $\ce$ stands for cross entropy. 
We will give more details on how $\ltok$ is defined for different inference network families below. It is also the loss used in our non-structured baseline models.

\subsection{Inference Network Architectures}

We now describe options for inference network architectures for sequence labeling. 
For each, we optionally include the modeling improvements described in Section~\ref{sec:strongerBaselines}. 
When doing so, we append ``+'' to the setting's  name to indicate this (e.g., $\infnetLarge$). 

\subsubsection{Convolutional Neural Networks} 
CNNs are frequently used in NLP to extract features based on symbol subsequences, whether words or characters~\citep{2011:NLP,P14-1062,DemnlpKim14,DBLP:journals/corr/KimJSR15,NIPS2015_5782}.  
CNNs use filters that are applied to symbol sequences and are typically followed by some sort of pooling operation. 
We apply filters over a fixed-size window centered on the word being labeled and do not use pooling. 
The feature maps $f_{n}(\x, t)$ for $(2n+1)$-gram filters are defined:
\vspace{-.08in}
\begin{align}
f_{n}(\x, t) = g(\w_n [\mathbf{v}_{x_{t-\window}};
...;\mathbf{v}_{x_{t+\window}}]+\mathbf{b}_n)  \nonumber
\end{align}
\noindent where $g$ is a nonlinearity, $\mathbf{v}_{x_{t}}$ is the embedding of word $x_t$, 
and $\w_n$ and $\mathbf{b}_n$ are filter parameters. 
We consider two CNN configurations: one 
uses $n=0$ and $n=1$ and the other 
uses $n=0$ and $n=2$. For each, we concatenate the two feature maps and use them as input to the softmax layer over outputs. 
In each case, we use $H$ filters for each feature map. 

\subsubsection{Recurrent Neural Networks}

For sequence labeling, it is common to use a BLSTM 
that runs over the input sequence and produces a softmax distribution over labels at each position in the sequence. We use this ``BLSTM tagger'' as our RNN inference network architecture. The parameter $H$ refers to the size of the hidden vectors in the forward and backward LSTMs, so the full dimensionality passed to the softmax layer is $2H$.

\subsubsection{Sequence-to-Sequence Models} 
Sequence-to-sequence (seq2seq; \citealt{sutskever2014sequence}) models have been successfully used for many sequential modeling tasks. 
It is common to augment models with an attention mechanism that focuses on particular positions of the input sequence while generating the output sequence~\citep{BahdanauCB14}. 
Since sequence labeling tasks have equal input and output sequence lengths and a strong connection between corresponding entries in the sequences, \citet{goyal18aaai} used fixed attention that deterministically attends to the $i$th input when decoding the $i$th output, and hence does not learn any attention parameters. It is shown as follows: 
\begin{align}
P(\mathbf{y}_t \mid \mathbf{y}_{<t},\x) &= \softmax(\w_s [\mathbf{h}_t, \mathbf{s}_t]) \nonumber
\end{align}
where $\mathbf{s}_t$ is the hidden vector at position $t$ from a BLSTM run over $\x$, $\mathbf{h}_t$ is the decoder hidden vector at position $t$, and $\w_s$ is a parameter matrix. The concatenation of the two hidden vectors is used to produce the distribution over labels. 

When using this inference network, we redefine the local loss to the standard training criterion for seq2seq models, namely the sum of the log losses for each output conditioned on the previous outputs in the sequence. 
We always use the previous predicted label as input (as used in ``scheduled sampling,''~\citealp{bengio2015scheduled}) during training because it works better for our tasks. 
In our experiments, the forward and backward encoder LSTMs use hidden dimension $H$, as does the LSTM decoder. 
Thus the model becomes similar to the BLSTM tagger except with conditioning on previous labeling decisions in a left-to-right manner.

We also experimented with the use of beam search for both the seq2seq baseline and inference networks and did not find much difference in the results. Also, as alternatives to the deterministic position-based attention described above, we experimented with learned local attention~\citep{LuongPM15} and global attention, but they did not work better on our tasks. 

\subsection{Methods to Improve Inference Networks} 
To further improve the performance of an inference network for a particular 
test instance $\x$, we propose two novel approaches that leverage the strengths of inference networks to provide effective starting points and then use instance-level fine-tuning in two different ways. 

\subsubsection{Instance-Tailored Inference Networks}

For each test example $\x$, we initialize an instance-specific inference network $\infnet(\x)$ using the trained inference network parameters, then run gradient descent on the following loss: 
\begin{align}
\argmin_{\Psi} E_\Theta(\x, \infnet(\x))
\label{eq:finetuneloss}
\end{align}
This procedure fine-tunes the inference network parameters for a single test example to minimize the energy of its output. For each test example, the process is repeated, with a new instance-specific inference network being initialized from the trained inference network parameters. 

\subsubsection{Warm-Starting Gradient Descent with Inference Networks}
Given a test example $\x$, we initialize ${\y}\in\relyspacer(\x)$ using the inference network and then use gradient descent by solving Eq.~\ref{eq:gdinf} described in Section~\ref{sec:gd} to update $\y$. 
However, the inference network output is in $\relyspace(\x)$ while gradient descent works with the more relaxed space $\relyspacer(\x)$. So we simply use the logits from the inference network, which are the score vectors before the softmax operations.

\section{Experimental Setup}
We perform experiments on three 
tasks: Twitter part-of-speech tagging (POS), named entity recognition (NER), and CCG supersense tagging (CCG). 

\subsection{Datasets}
\paragraph{POS.}

We use the annotated data from \citet{gimpel-11a} and \citet{owoputi-EtAl:2013:NAACL-HLT} which contains
25 POS tags. For training, we combine the 1000-tweet \textsc{Oct27Train} set and the 327-tweet \textsc{Oct27Dev} set. For validation, we use the 500-tweet \textsc{Oct27Test} set and for testing we use the 547-tweet \textsc{Daily547} test set. We use the 100-dimensional skip-gram embeddings from \citet{tu-17-long} which were trained on a dataset of 56 million English tweets using  \texttt{word2vec}~\citep{mikolov2013distributed}. The evaluation metric is tagging accuracy.

\vspace{-0.2cm}
\paragraph{NER.}
We use the CoNLL 2003 English data \citep{TjongKimSang:2003:ICS:1119176.1119195}. 
There are four entity types: PER, LOC, ORG, and MISC. 
There is a strong local dependency between neighboring labels because this is a labeled segmentation task. 
We use the BIOES tagging scheme, so there are 17 labels. 
We use 100-dimensional pretrained GloVe~\citep{pennington2014glove} embeddings.  
The task is evaluated with micro-averaged F1 score using the \texttt{conlleval} script. 

\vspace{-0.2cm}
\paragraph{CCG.}
We use the standard splits from CCGbank~\citep{HockenmaierS02}. 
We only keep sentences with length less than 50 in the original training data when training the CRF. 
The training data contains 1,284 unique labels, but 
because the label distribution has a long tail, 
we use only the 400 most frequent labels, replacing the others by a special tag $*$. 
The percentages of $*$ in train/development/test are 0.25/0.23/0.23$\%$. 
When the gold standard tag is $*$, the prediction is always evaluated as incorrect. 

We use the same GloVe embeddings as in NER. 
Because of the compositional nature of supertags, this task has more non-local dependencies. 
The task is evaluated with per-token accuracy.
\begin{table*}[t]
\centering
\small
\begin{tabular}{l|c|c|c||c|c|c||c|c|c|}
\cline{2-10}
& \multicolumn{3}{c||}{Twitter POS Tagging}
& \multicolumn{3}{c||}{NER}
& \multicolumn{3}{c|}{CCG Supertagging}
\\ \cline{2-10} 
& CNN & BLSTM & seq2seq & CNN & BLSTM & seq2seq &CNN & BLSTM & seq2seq \\\hline
\multicolumn{1}{|l|}{local baseline} & 89.6 & 88.0 & 88.9 & 79.9 & 85.0 & 85.3 & 90.6& 92.2  &  92.7\\
\hline
\multicolumn{1}{|l|}{$\infnetBase$} & \bf 89.9 & 89.5  & 89.7   & 82.2 & 85.4  & 86.1  & 91.3 & 92.8  & \bf 92.9  \\
\hline
\multicolumn{1}{|l|}{gradient descent} & \multicolumn{3}{c||}{89.1} & \multicolumn{3}{c||}{84.4} & \multicolumn{3}{c|}{89.0} \\
\hline
\multicolumn{1}{|l|}{Viterbi } & \multicolumn{3}{c||}{89.2} & \multicolumn{3}{c||}{\bf 87.2} & \multicolumn{3}{c|}{92.4} \\
\hline


\end{tabular}
\caption{Test results for all tasks. Inference networks, gradient descent, and Viterbi are all optimizing the \blstmcrf energy. Best result per task is in bold.
}
\label{table:loss}
\end{table*}

\subsection{Training and Tuning}

For the optimization problems mentioned below, we use stochastic gradient descent with momentum as the optimizer. Full details of hyperparameter tuning are in the appendix. 

\paragraph{Local Baselines.} 
We consider local (non-structured) baselines that use the same architectures as the inference networks but train using only the local loss $\ltok$. 

\paragraph{Structured Baselines.} 
We train the \blstmcrf and \blstmcnncrf  models with the standard conditional log-likelihood objective. We tune hyperparameters on the development sets. 

\paragraph{Gradient Descent for Inference.} 

We use gradient descent for structured inference by solving Eq.~\ref{eq:gdinf}. 
We randomly initialize ${\y}\in\relyspacer(\x)$ and, for $N$ iterations, we compute the gradient of the energy with respect to $\y$, 
then update $\y$ using gradient descent with momentum, which we found to generally work better than constant step size. 
We tune $N$ and the learning rate via instance-specific oracle tuning, i.e., we choose them separately for each input to maximize performance (accuracy or F1 score) on that input. Even with this oracle tuning, we find that gradient descent struggles to compete with the other methods. 


\paragraph{Inference Networks.} 
To train the inference networks, we first 
train the \blstmcrf or \blstmcnncrf model with the standard conditional log-likelihood objective. The hidden sizes $H$ are tuned in that step. 
We then fix the energy function and train the inference network $\infnet$ using the combined loss from Section~\ref{sec:infnet}. 

For instance-tailored inference networks and when using inference networks as a warm start for gradient descent, we tune the number of epochs $N$ and the learning rate on the development set, and report the performance on the test set, using the same values of $N$ and the learning rate for all test examples.

\section{\blstmcrf Results}

This first section of results uses the simpler \blstmcrf modeling configuration. In Section~\ref{sec:res-larger} below we present results with the stronger \blstmcnncrf configuration and also apply the same modeling improvements to the baselines and inference networks. 


Table~\ref{table:loss} shows test results for all tasks and architectures. 
The inference networks use the same architectures as the corresponding local baselines, but their parameters are trained with both the local loss and the \blstmcrf energy, leading to consistent improvements. 
CNN inference networks work well for POS, but struggle on NER and CCG compared to other architectures. 
BLSTMs work well, but are outperformed slightly by seq2seq models across all three tasks. 
Using the Viterbi algorithm for exact inference yields the best performance for NER but is not best for the other two tasks. 

It may be surprising that an inference network trained to mimic Viterbi would outperform Viterbi in terms of accuracy, which we find for the CNN for POS tagging and the seq2seq inference network for CCG. We suspect this occurs for two reasons. One is due to the addition of the local loss in the inference network objective; the inference networks may be benefiting from this multi-task training. 
\citet{structure-loss} similarly found benefit from a combination of token-level and sequence-level losses. 
The other potential reason is beneficial inductive bias with the inference network architecture. For POS tagging, the CNN architecture is clearly well-suited to this task given the strong performance of the local CNN baseline. Nonetheless, the CNN inference network is able to improve upon both the CNN baseline and Viterbi. 

\begin{figure}[t]
  \centering
  \begin{subfigure}[b]{0.49\linewidth}
    \includegraphics[width=\linewidth]{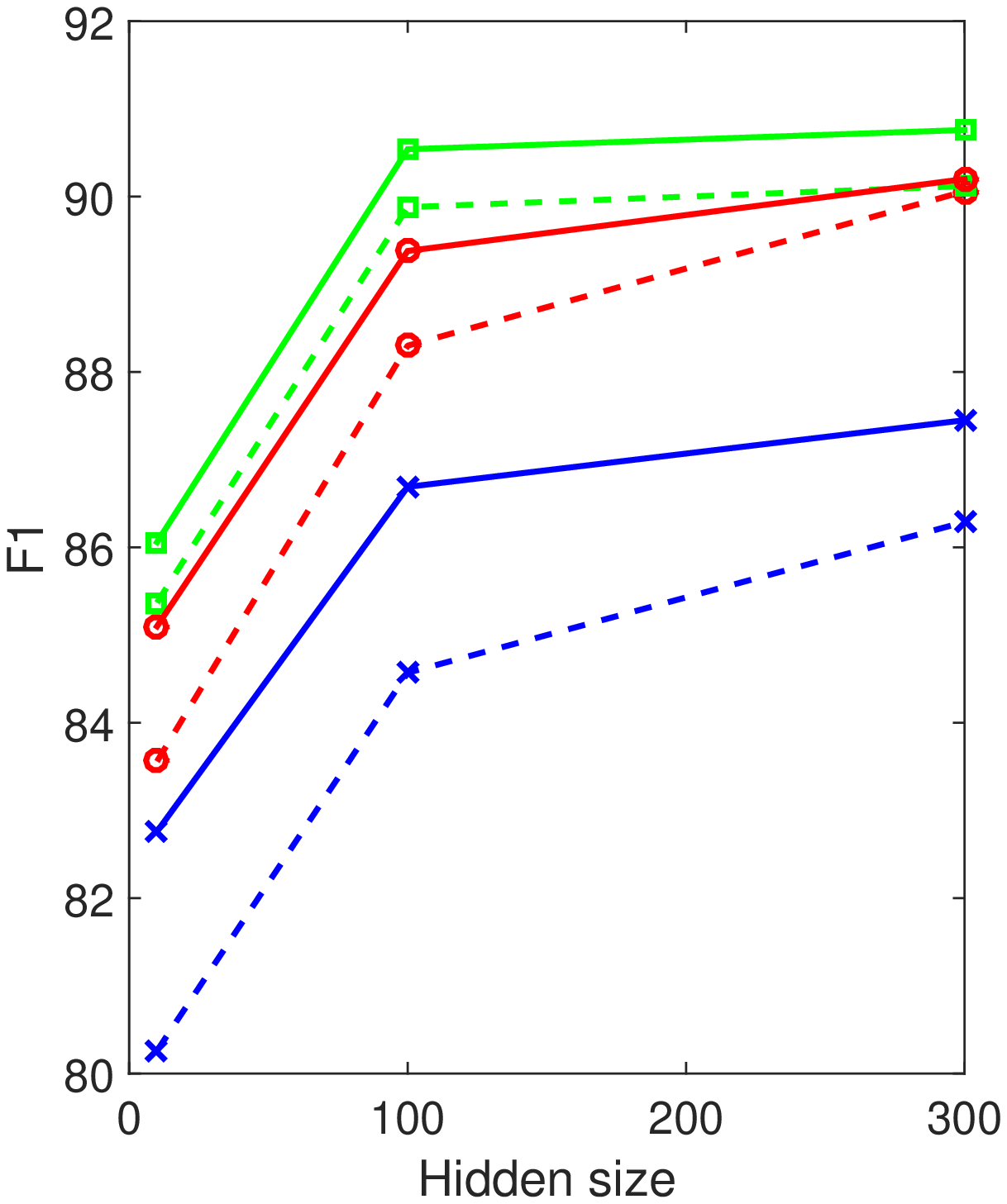}
    \caption{NER}
  \end{subfigure}
  \begin{subfigure}[b]{0.49\linewidth}
    \includegraphics[width=\linewidth]{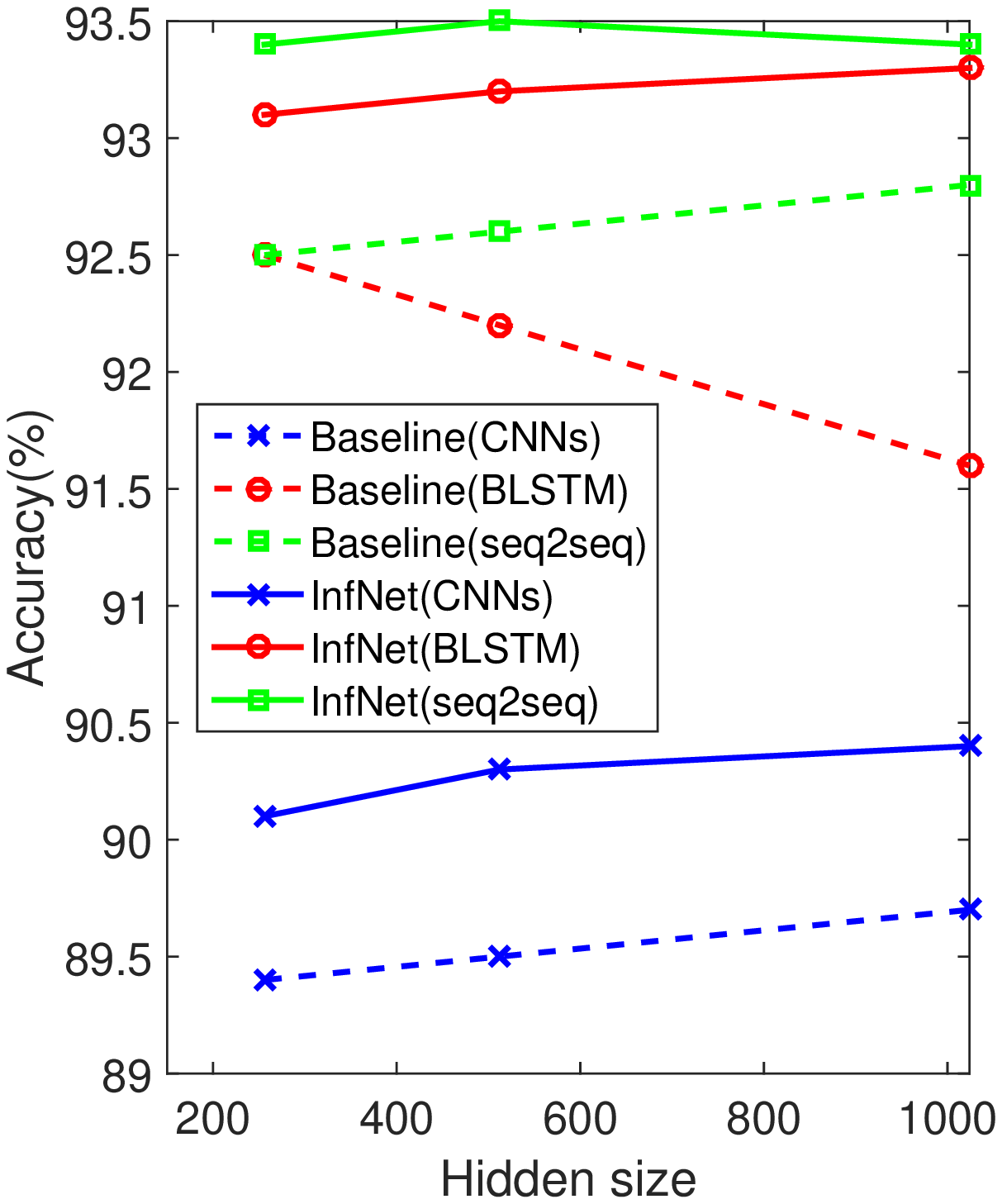}
    \caption{CCG Supertagging}
  \end{subfigure}
  \caption{Development results for inference networks with different architectures and hidden sizes ($H$).
  }
  \label{fig:hiddensize}
\end{figure}

\paragraph{Hidden Sizes.} For the test results in Table~\ref{table:loss}, we did limited tuning of $H$ for the inference networks based on the development sets. Figure~\ref{fig:hiddensize} shows the impact of $H$ on performance. Across $H$ values, the inference networks outperform the baselines. For NER and CCG, seq2seq outperforms the BLSTM which in turn outperforms the CNN.

\begin{table}[t]
\centering
\small
\begin{tabular}{|ll|c|c|}
\cline{3-4}
\multicolumn{2}{l|}{} & $\{$1,3$\}$-gram & $\{$1,5$\}$-gram\\
\hline
\multirow{2}{*}{POS} & local baseline & 89.2&  88.7\\ 
 & $\infnetBase$ & \bf 89.6 & 89.0\\ 
\hline
\multirow{2}{*}{NER} & local baseline &84.6   &  85.4\\ 
 & $\infnetBase$  & 86.7  & \bf 86.8\\
\hline
\multirow{2}{*}{CCG} & local baseline & 89.5  & 90.4\\
 & $\infnetBase$  &  90.3 & \bf 91.4\\
\hline
\end{tabular}
\caption{Development results for CNNs with two filter sets ($H=100$).}
\label{table:cnnresults}
\end{table}

\paragraph{Tasks and Window Sizes.} 
Table~\ref{table:cnnresults} shows that CNNs with smaller windows are better for POS, while larger windows are better for NER and CCG. This suggests that POS has more local dependencies among labels than NER and CCG. 

\subsection{Speed Comparison}
Asymptotically, Viterbi takes $\mathcal{O}(nL^2)$ time, where 
$n$ is the sequence length. The BLSTM and our deterministic-attention seq2seq models have time complexity $\mathcal{O}(nL)$. 
CNNs 
also have complexity $\mathcal{O}(nL)$ but are more easily parallelizable. 
Table~\ref{table:speed} shows test-time inference speeds for inference networks, gradient descent, and Viterbi for the \blstmcrf model. We use GPUs and a minibatch size of 10 for all methods. 
CNNs are 1-2 orders of magnitude faster than the others. BLSTMs work almost as well as seq2seq models and are 2-4 times faster in our experiments. 

Viterbi is actually faster than seq2seq when $L$ is small, but for CCG, which has $L=400$, it is 4-5 times slower. Gradient descent is slower than the others because it generally needs many iterations (20-50) for competitive performance. 

\begin{table}[t]
\setlength{\tabcolsep}{4pt}
\centering
\small
\begin{tabular}{|l|c|c|c|c|c|}
\cline{2-6}
\multicolumn{1}{l|}{} & CNN & BLSTM  & seq2seq & Viterbi & \gd
\\ \hline
POS & 12500 & 1250  & 357   & 500 & 20 \\  

NER &  10000 & 1000  & 294 & 360 & 23\\

CCG & 6666 & 1923 &  1000 & 232 & 16 \\

\hline
\end{tabular}
\caption{Speed comparison of inference networks across tasks and architectures (examples/sec). 
}
\label{table:speed}
\end{table}

\subsection{Search Error} 
We can view inference networks as approximate search algorithms and assess characteristics that affect  search error. 
To do so, we train two 
LSTM language models (one on word sequences and one on gold label sequences) on the Twitter POS data. 
We also compute the difference in the \blstmcrf energies between the inference network output $\y_{\info}$ and the Viterbi output $\y_{\vito}$ as the search error: 
$E_{\Theta}(\x, \y_{\info}) - E_{\Theta}(\x, \y_{\vito})$. 
We compute the same search error for gradient descent.

For the BLSTM inference network, Spearman's $\rho$ between the word sequence perplexity and search error is 0.282; for the label sequence perplexity, it is 0.195.
For gradient descent inference, Spearman's $\rho$ between the word sequence perplexity and search error is 0.122; for the label sequence perplexity, it is 0.064.
These positive correlations mean that for frequent sequences, inference networks and gradient descent exhibit less search error. We also note that the correlations are higher for the inference network than for gradient descent, showing the impact of amortization during learning of the inference network parameters. That is, since we are learning to do inference from a dataset, we would expect search error to be smaller for more frequent sequences, and we do indeed see this correlation.

\section{\blstmcnncrf Results}
\label{sec:res-larger}


We now compare inference methods when using the improved modeling techniques described in Section~\ref{sec:strongerBaselines} (i.e., the setting we called \blstmcnncrf). We use these improved techniques for all models, including the CRF, the local baselines, gradient descent, and the inference networks. When training inference networks, both the inference network 
architectures and the structured energies use the techniques from Section~\ref{sec:strongerBaselines}. So, when referring to inference networks in this section, we use the name $\infnetLarge$. 

The results are shown in Table~\ref{table:advanceloss}. 
With a more powerful local architecture, structured prediction is less helpful overall, but 
inference networks still improve over the local baselines 
on 2 of 3 tasks. 
\paragraph{POS.}
As in the \blstmcrf setting, the local CNN baseline and the CNN inference network outperform Viterbi. This is likely because the CRFs use BLSTMs as feature networks, but our results show that CNN baselines are consistently better than BLSTM baselines on this task.
 
As in the \blstmcrf setting, gradient descent works quite well on this task, comparable to Viterbi, though it is still much slower.

\begin{table}[t]
\centering
\small
\begin{tabular}{|l|c|c|c|}
\cline{2-4}
\multicolumn{1}{l|}{} &POS
& NER
& CCG \\
\hline
local baseline & 91.3 & 90.5 & 94.1 \\ 
$\infnetLarge$  & 91.3 & 90.8 & 94.2 \\
gradient descent & 90.8 & 89.8 & 90.4 \\
Viterbi & 90.9 & 91.6 & 94.3 \\ \hline
\end{tabular}
\caption{Test results 
with \blstmcnncrf. For local baseline and inference network architectures, we use 
CNN for POS, seq2seq for NER, and BLSTM for CCG.
}
\label{table:advanceloss}
\end{table}

\begin{table}[t]
\centering
\small
\begin{tabular}{|l|c|}
\cline{2-2}
\multicolumn{1}{l|}{}  & F1 
\\ \hline
local baseline (BLSTM)  & 90.3 \\ 
$\infnetLarge$ (1-layer BLSTM) & 90.7 \\
$\infnetLarge$ (2-layer BLSTM) & 91.1 \\
Viterbi &  91.6 \\
\hline
\end{tabular}
\caption{NER test results (for \blstmcnncrf)  with more layers 
in the 
BLSTM inference network.} 
\label{table:morelayer_ner_result}
\end{table}

\begin{table*}[t]
\centering
\small
\begin{tabular}{lc|c|c||c|c||c|c|}
\cline{3-8}
&& \multicolumn{2}{c||}{Twitter POS Tagging}
& \multicolumn{2}{c||}{NER}
& \multicolumn{2}{c|}{CCG Supertagging}
\\ \cline{3-8} 
& $N$ & Acc. $(\uparrow)$ & Energy $(\downarrow)$   & F1 $(\uparrow)$ & Energy $(\downarrow)$ & Acc. $(\uparrow)$ & Energy  $(\downarrow)$ \\
\hline
\multicolumn{1}{|l}{gold standard} &  & 100 & -159.65 &100    & -230.63 &  100 &  -480.07  \\
\hline \hline
\multicolumn{1}{|l}{\blstmcnncrf/Viterbi} &  & 90.9  & -163.20 & 91.6   & -231.53 & 94.3  & -483.09   \\
\hline \hline
\multicolumn{1}{|l}{} & 10 & 89.2 & -161.69 &81.9 & -227.92 & 65.1 &-412.81  \\
\multicolumn{1}{|l}{} & 20 &90.8  & -163.06 &  89.1  & -231.17 &  74.6 & -414.81   \\
\multicolumn{1}{|l}{} & 30 & 90.8 & -163.02 & 89.6 & -231.30 & 83.0 & -447.64\\
\multicolumn{1}{|l}{\multirow{2}{*}{gradient descent}} & 40 & 90.7 & -163.03 &  89.8  & -231.34 & 88.6  &-471.52    \\
\multicolumn{1}{|l}{} & 50 & 90.8 & -163.04 &  89.8  & -231.35 &  90.0 &-476.56    \\
\multicolumn{1}{|l}{} & 100 & - & - &  -  & - &  90.1 &-476.98    \\
\multicolumn{1}{|l}{} & 500 & - & - & -  & - &  90.1 &-476.99    \\
\multicolumn{1}{|l}{} & 1000 &- & - &  -  & - &  90.1 &-476.99    \\
\hline \hline
\multicolumn{2}{|l|}{$\infnetLarge$} & 91.3 & -162.07 &  90.8  & -231.19 & 94.2 & -481.32    \\
\multicolumn{2}{|l|}{discretized output from $\infnetLarge$} & 91.3 & -160.87 & 90.8   & -231.34 & 94.2  & -481.95    \\
\hline \hline
\multicolumn{1}{|l}{} & 3 &91.0  &-162.59  &91.3  &-231.32  & 94.3 & -481.91    \\
\multicolumn{1}{|l}{instance-tailored $\infnetLarge$} & 5 & 90.9 & -162.81 & 91.2 & -231.37 & 94.3 &  -482.23   \\
\multicolumn{1}{|l}{} & 10 & 91.3 & -162.85 & 91.5 & -231.39 & 94.3 & -482.56    \\
\hline \hline
\multicolumn{1}{|l}{\multirow{2}{*}{$\infnetLarge$ as warm start for}} & 3 &91.4  &-163.06  & 91.4 & -231.42 & 94.4 &-482.62     \\
\multicolumn{1}{|l}{\multirow{2}{*}{gradient descent}} & 5 & 91.2 & -163.12  & 91.4 & -231.45 & 94.4 & -482.64    \\
\multicolumn{1}{|l}{} & 10 &91.2  & -163.15 & 91.5 & -231.46 &  94.4& -482.78    \\
\hline
\end{tabular}
\vspace{-0.1cm}
\caption{Test set results of approximate inference methods for three tasks, showing performance metrics (accuracy and F1) as well as average energy of the output of each method. 
The inference network architectures in the above experiments are: CNN for POS, seq2seq for NER, and BLSTM for CCG. $N$ is the number of epochs for GD inference or instance-tailored fine-tuning. 
}
\label{table:sgd}
\vspace{-0.2cm}
\end{table*}

\paragraph{NER.} 
We see slightly higher \blstmcnncrf results than several previous state-of-the-art results (cf.~90.94;~\citealp{LampleBSKD16} and 91.37;~\citealp{ma-hovy:2016:P16-1}). 
The stronger \blstmcnncrf configuration also helps the inference networks, 
improving performance from 90.5 to 90.8 for the seq2seq architecture over the local baseline. 
Though gradient descent reached high accuracies for POS tagging, it does not perform well on NER, possibly due to the greater amount of non-local information in the task. 

While we see strong performance with \infnetLarge, it still lags behind Viterbi in F1. 
We consider additional experiments in which we increase the number of layers in the inference networks. We use a 2-layer BLSTM as the inference network and also use weight annealing of the local loss hyperparameter $\lambda$, setting it to $\lambda=e^{-0.01t}$ where $t$ is the epoch number. Without this annealing, the 2-layer inference network was difficult to train. 
The weight annealing was helpful for encouraging the inference network to focus more on the non-local information in the energy function rather than the token-level loss. 
As shown in Table~\ref{table:morelayer_ner_result}, these changes yield an improvement of 0.4 in F1. 

\paragraph{CCG.} 
Our \blstmcnncrf reaches an accuracy of 94.3\%, which is comparable to several recent results (93.53, \citealp{XuAC16}; 94.3, \citealp{N16-1026}; and 94.50,  \citealp{vaswani}). 
The local baseline, the BLSTM inference network, and Viterbi are all extremely close in accuracy. 
Gradient descent struggles here, likely due to the large number of candidate output labels.


\subsection{Speed, Accuracy, and Search Error}
Table~\ref{table:sgd} compares inference methods in terms of both accuracy and energies reached during inference. 
For each number $N$ of gradient descent iterations in the table, we tune the learning rate per-sentence and report the average accuracy/F1 with that fixed number of iterations. We also report the average energy reached. 
For inference networks, we report energies both for the output directly and when we discretize the output (i.e., choose the most probable label at each position). 

\paragraph{Gradient Descent Across Tasks.} 
The number of gradient descent iterations required for competitive performance varies by task. For POS, 20 iterations are sufficient to reach accuracy and energy close to Viterbi. 
For NER, roughly 40 iterations are needed for gradient descent to reach its highest F1 score, and for its energy to become very close to that of the Viterbi outputs. However, its F1 score is much lower than Viterbi. 
For CCG, gradient descent requires far more iterations, presumably due to the larger number of labels in the task. Even with 1000 iterations, the accuracy is 4\% lower than Viterbi and the inference networks. Unlike POS and NER, the inference network reaches much lower energies than gradient descent on CCG, suggesting that the inference network may not suffer from the same challenges of searching high-dimensional label spaces as those faced by gradient descent. 

\paragraph{Inference Networks Across Tasks.}
For POS, the inference network does not have lower energy than gradient descent with $\geq 20$ iterations, but it does have higher accuracy. 
This may be due in part to our use of multi-task learning for inference networks. 

The discretization of the inference network outputs increases the energy on average for this task, whereas it decreases the energy for the other two tasks. 
For NER, the inference network reaches a similar energy as gradient descent, especially when discretizing the output, but is considerably better in F1. 
The CCG tasks shows the largest difference between gradient descent and the inference network, as the latter is much better in both accuracy and energy. 

\paragraph{Instance Tailoring and Warm Starting.}
Across tasks, instance tailoring and warm starting lead to lower energies than $\infnetLarge$. The improvements in energy are sometimes joined by improvements in accuracy, notably for NER where the gains range from 0.4 to 0.7 in F1. 
Warm starting 
gradient descent yields the lowest energies (other than Viterbi), showing promise for the use of gradient descent as a local search method starting from inference network output.

\begin{figure}[t]
  \centering
\includegraphics[width=\linewidth]{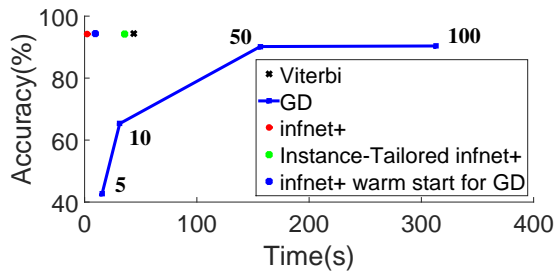}
  \caption{CCG test results for inference methods (GD = gradient descent). 
  The x-axis is the total inference time for the test set. The numbers on the GD curve are the number of gradient descent iterations. 
  }
  \label{fig:inference_method}
\end{figure}

\paragraph{Wall Clock Time Comparison.} 
Figure \ref{fig:inference_method} shows the speed/accuracy trade-off for the inference methods, using wall clock time for test set inference as the speed metric. 
On this task, Viterbi is time-consuming because of the larger label set size. The inference network has comparable accuracy to Viterbi but is much faster. Gradient descent needs much more time to get close to the others but plateaus before actually reaching similar accuracy. Instance-tailoring and warm starting reside between $\infnetLarge$ and Viterbi, with warm starting being significantly faster because it does not require updating inference network parameters.

\section{Related Work}

%
The most closely related prior work is that of \citet{tu-18}, who 
experimented with RNN inference networks for sequence labeling. We compared three architectural families, showed the relationship between optimal architectures and downstream tasks, compared inference networks to gradient descent, and proposed novel variations. 

We focused in this paper on sequence labeling, in which CRFs with neural network potentials have emerged as a state-of-the-art approach~\citep{LampleBSKD16,ma-hovy:2016:P16-1,StrubellVBM17,C18-1327}. 
Our results suggest that inference networks can provide a feasible way to speed up test-time inference over Viterbi without much loss in performance. 
The benefits of inference networks may be coming in part from multi-task training; 
\citet{structure-loss} 
similarly found benefit from combining token-level and sequence-level losses. 


We focused on structured prediction in this paper, but inference networks are useful in other settings as well. For example, it is common to use a particular type of inference network to approximate posterior inference in neural approaches to latent-variable probabilistic modeling, such as variational autoencoders~\citep{KingmaW13} and, more closely related to this paper, variational sequential labelers~\citep{D18-1020}. In such settings, \citet{pmlr-v80-kim18e} have found benefit with instance-specific updating of inference network parameters, which is related to our instance-level fine-tuning. 
There are also connections between structured inference networks and amortized structured inference \citep{Srikumar:2012:AIC:2390948.2391073} as well as methods for neural knowledge distillation and model compression \citep{hinton2015distilling,NIPS2014_5484,D16-1139}. 

Gradient descent is used for inference in several settings, e.g., structured prediction energy networks \citep{belanger2016structured}, image generation applications~\citep{deedream,DBLP:GatysEB15a},  
finding adversarial examples \citep{43405}, 
learning paragraph embeddings \citep{Le:2014:DRS:3044805.3045025}, 
and machine translation \citep{DBLP:conf/emnlp/HoangHC17}.
Gradient descent has started to be replaced by inference networks in some of these settings, such as 
image transformation  \citep{Johnson2016Perceptual,DBLP:journals/corr/LiW16b}. Our results provide more evidence that gradient descent can be replaced by inference networks or improved through combination with them. 


\section{Conclusion}
We compared several methods for approximate inference in neural structured prediction, finding that inference networks achieve a better speed/accuracy/search error trade-off than gradient descent. 
We also proposed instance-level inference network fine-tuning and using inference networks to initialize gradient descent, finding further 
reductions in search error and improvements in performance metrics for certain tasks.

\section*{Acknowledgments}
We would like to thank Ke Li for suggesting experiments that combine inference networks and gradient descent, the anonymous reviewers for their feedback, 
and NVIDIA for donating GPUs used in this research. 

\bibliography{naaclhlt2019}
\bibliographystyle{acl_natbib}

\appendix


\section{Appendix} 

\label{sec:hypers}
\paragraph{Local Baselines.} 
We consider local (non-structured) baselines that use the same architectures as the inference networks but train using only the local loss $\ltok$. We tune the learning rate ($\{5,1,0.5, 0.1, 0.05, 0.01, 0.005, 0.001, 0.0005\}$). We train on the training set, use the development sets for tuning and early stopping, and report results on the test sets. 

\paragraph{Structured Baselines.} 
We train the \blstmcrf and \blstmcnncrf  models with the standard conditional log-likelihood objective. We tune hyperparameters on the development sets. The tuned BLSTM hidden size $H$ for \blstmcrf is 100 for POS/NER and 512 for CCG; for \blstmcnncrf the tuned hidden size is 100 for POS, 200 for NER, and 400 for CCG. 

\paragraph{Gradient Descent for Inference.} 

For the number of epochs $N$, we consider values in the set $\{5, 10, 20, 30, 40, 50, 100, 500, 1000\}$. For each $N$, we tune the learning rate over the set  $\{1\mathrm{e}^4,5\mathrm{e}^3,1\mathrm{e}^3, 500, 100, 50, 10, 5, 1\}$). These learning rates may appear extremely large when we are accustomed to choosing rates for empirical risk minimization, but we generally found that the most effective learning rates for structured inference are orders of magnitude larger than those effective for learning. 
To provide as strong performance as possible for the gradient descent method, we tune $N$ and the learning rate via oracle tuning, i.e., we choose them separately for each input to maximize performance (accuracy or F1 score) on that input.

\paragraph{Inference Networks.} 
To train the inference networks, we first 
train the \blstmcrf or \blstmcnncrf model with the standard conditional log-likelihood objective. The hidden sizes $H$ are tuned in that step. 
We then fix the energy function and train the inference network $\infnet$ using the combined loss from Section 4. We tune the learning rate over the set $\{5,1,0.5, 0.1, 0.05, 0.01, 0.005, 0.001, 0.0005\}$ for the inference network and the local loss weight $\lambda$ over the set $\{0.2, 0.5,1,2,5\}$.  
We use early stopping on the development sets and report the results on the test sets using the trained inference networks. 

\end{document}